\documentclass[letterpaper, 10 pt, conference]{ieeeconf}  

\IEEEoverridecommandlockouts                              

\usepackage{tabu}
\usepackage[utf8]{inputenc}
\usepackage{graphicx}
\usepackage[table,xcdraw]{xcolor}
\usepackage[export]{adjustbox}
\usepackage{tabularx}
\usepackage{array}
\usepackage{hyperref}
\usepackage{balance}
\usepackage{lipsum}
\usepackage{float}
\usepackage{multirow}
\usepackage{color}

\usepackage{subcaption}
\usepackage{booktabs}
\usepackage{multirow}

\newcolumntype{P}[1]{>{\centering\arraybackslash}m{0.04\linewidth}}

\hypersetup{
  pdfauthor={Gustavo Assuncao, B.Sc.},
  pdftitle={TCEDI},
  pdfsubject={},
    pdfkeywords={},
    pdfproducer={Latex with hyperref},
    pdfcreator={overleaf}]{hyperref}
}

\title{\LARGE \bf Approaching Metaheuristic Deep Learning Combos for Automated Data Mining}

\author{Gustavo Assunção$^{1,2}$, Paulo Menezes$^{1,2}$
\thanks{$^{1}$Institute of Systems and Robotics,  University of Coimbra, R. Silvio Lima, 3030-194 Coimbra, Portugal {\tt \{gustavo.assuncao@isr.uc.pt, pm@deec.uc.pt\}}}%
\thanks{$^{2}$Department of Electrical and Computer Engineering, University of Coimbra, R. Silvio Lima, 3030-194 Coimbra, Portugal}
}

\begin{document}

\maketitle
\thispagestyle{empty}
\pagestyle{empty}

\begin{abstract}

Lack of data on which to perform experimentation is a recurring issue in many areas of research, particularly in machine learning. The inability of most automated data mining techniques to be generalized to all types of data is inherently related with their dependency on those types which deems them ineffective against anything slightly different. Meta-heuristics are algorithms which attempt to optimize some solution independently of the type of data used, whilst classifiers or neural networks focus on feature extrapolation and dimensionality reduction to fit some model onto data arranged in a particular way. These two algorithmic fields encompass a group of characteristics which when combined are seemingly capable of achieving data mining regardless of how it is arranged. To this end, this work proposes a means of combining meta-heuristic methods with conventional classifiers and neural networks in order to perform automated data mining. Experiments on the MNIST dataset for handwritten digit recognition were performed and it was empirically observed that using a ground truth labeled dataset's validation accuracy is inadequate for correcting labels of other previously unseen data instances.

\end{abstract}

\section{Introduction}
    
    Given current research trends and thanks to technological advancements in parallel and high performance computing, the application of machine learning and neural networks in data science has become the standard for success. With that in mind, the vast majority of automated learning approaches has become increasingly dependent on the availability of staggeringly large good-quality datasets, colloquially known as big data \cite{vial2020}. Accordingly, this may entail different degrees of complexity correlated with the type of data considered. In addition, labelling raw data requires a great deal of human effort and often specialty training when dealing with potential intricacies and ramifications. Furthermore, human error is unavoidable and its propagation must always be accounted for, given it may undermine the validity of seemingly successful results \cite{Barchard_2011}.
    
    Artificial Neural networks (ANNs) can be interpreted as structures whose architecture can be adjusted to fit specific types of data, optimizing their analysis through the learning of correlations between extracted features. As such, these are typically employed in supervised learning scenarios aimed at the subsequent prediction of new labels for previously unseen data. Given the variability associated with a network's many hyperparameters and the lack of methodology to obtain optimal combinations, their performance must always be empirically improved and perfect accuracy may not be reached. Moreover, this also leads to error propagation should the mere predictions of an ANN be taken as completely true. Thus, artificial neural networks may serve as evaluation methods for novel data \cite{ghazal2020data}, yet still lack the autonomy to function as standalone systems \cite{mythesis}. Ways to attain or, in some form, boost this skill include synergizing networks with other systems towards more holistic problem solving. For instance, bio-inspired systems can serve as an add-on when developing more autonomous neural network approaches, as in \cite{Assun_o_2021}. Heuristics are another type of methodology which can also help mitigate some gap in network capabilities.

    Meta-heuristics, procedures which empirically strive for optimization without providing theoretical guarantees that a problem's optimal solution can be reached or even neared, are great candidates for dealing with the exploding combinatorial nature of data mining problems and neural network methodology. More specifically, genetic algorithms (GA) \cite{meth_2} are highly adequate options given the observed efficacy of the natural selection and genetic principles they attempt to mimic. As a result of apt individuals generating offspring which maintain their positive characteristics while also possibly obtaining new ones, GAs allow for a more efficient search of viable solutions. This can be applied in data mining, as a means to improve sets of automatically generated labels at each iteration of an algorithm. Another option is simulated annealing \cite{meth_3}, a meta-heuristic inspired by the physical process of heating a solid in order to obtain new low energy states for its condensed matter, by means of cooling it towards thermal equilibrium. The main difference between the two former meta-heuristics is that GAs are populational methods, considering several possible solutions at each iteration, whilst simulated annealing assesses the fitness of a single solution at each iteration and then moves to a new one at the following iteration.
    
    In this work we propose a novel combination of artificial neural networks within a meta-heuristic, designed to perform automated data mining. The system takes as input a small set of labelled instances as well as a larger set of raw data, for which it attempts to maximize the amount of correct generated labels. Two distinct heuristics are considered. Specifically, a GA is compared to a homologous simulated annealing one for performance evaluation. This way we address the lack of big data in most machine learning problems, allowing for the automatic production of large datasets while also reducing the need for manual labelling and time consumption.
    
    This paper is organized in the following manner. To start, an analysis of related state-of-the-art work is provided in Section II, so as to provide the user with some context on the coupling of neural networks and meta-heuristics. The proposed technique is described subsequently in Section III, followed by Section IV where the experimental testing portion of the work is explained. The obtained results from those experiments are shown in Section V, along with a corresponding discussion and critique. Finally, a conclusion is provided in Section VI.

\section{Related work}

    This overview of related work is divided in two parts. First, we present a short overview of recent systems used in automated data mining. Following that, approaches are analyzed whose basis is on the coupling of neural networks and meta-heuristics as a means to improve a goal ANNs alone are unable to.
    
    \subsection{Automated Data Mining}
    Due to the broad spectrum of types of data in the real world, data mining automated systems tend to be highly specific to each situation, not allowing for a generalization. Examples of these methods are used in \cite{rw_1_1} for automatic creation of a speaker recognition database, in \cite{rw_1_2} for visual speech data mining, or in \cite{10090765} for pooling industry logistics information, to give a few examples. Though they performed well in those works, there is little interest in attempting to extrapolate them to other areas of research, given their specificity. In any case, \cite{6724149} provides a short overview of machine learning techniques for data mining, which may be useful as inspiration for other specific developments.
    
    The low generalization problem is less overt in data mining systems based on heuristics, where techniques are included which do not necessarily rely on specific types of data to function properly. Though usually these boast lower performances, their greater generalizability is highly advantageous as it enables applicability to a growing diversity of data types and problems, typical of current artificial intelligence research \cite{Xu_2021}. For instance, in \cite{rw_1_3} Purushothama \textit{et al.} introduced the use of a genetic approach for mining closed sequential patterns in big data. Mohamadi \textit{et al.} \cite{rw_1_4} presented a technique in which simulated annealing was employed in extracting an optimal set of fuzzy classification rules from input datasets, which were then used for classifying other datasets. Both techniques were able to deal with different datasets, and hence, data heterogeneity.
    
    \subsection{Heuristic ANN Coupling}
    The combination of heuristic methods with neural networks been executed in several ways, mainly due to the fact that one side's benefits are able to compensate where the other falls short. In \cite{rw_2_1}, authors used the accuracy of an ANN as a fitness value for a GA in order to evolve a set of selected features for training and consequently better predict a stock index. Liang \textit{et al.} \cite{rw_2_2}, used a GA to adapt neuron weight vectors, optimizing a neural network to find better symbolic rules for data mining. The approach was tested with the IRIS dataset. Relatedly, Chen \textit{et al.} \cite{rw_2_3} employed a GA to find optimal initialization weights and biases for the training of a neural network. This allowed for the reduction of computational time and better prediction of a river's water-level. Heuristics have also been used to improve ANN architectures without a need for trial and error. In \cite{rw_2_4}, authors compared the use of genetic algorithmics, Taguchi methodology, tabu search and decision trees in finding the best hyperparameter combo (e.g. number of hidden layers) for ANNs used in manufacturing processes. Additionally, heuristics have been used for the direct optimization of ANN layers' weights and corresponding biases. Nimbark \textit{et al.} \cite{rw_2_5}, employed an artificial bee colony algorithm (meta-heuristic based on bees' food searching behavior) in finding optimal synaptic weights and transfer functions for ANN training. In \cite{rw_2_6}, an original heuristic neural network was developed for pattern recognition, using custom adjusting algorithms for improving weights and biases in each layer at each training iteration. In another approach \cite{rw_2_7}, authors implemented their own heuristic mutation operator to evolve network weights and structure (node addition/deletion) simultaneously, for higher accuracy with less computational time.
    
    In comparison with the approaches mentioned above, the proposed technique aims to combine a meta-heuristic method with an artificial neural network for automated data mining, which is in every sense independent of the considered data type. Unlike related techniques, this entails greater generalizability to unforeseen scenarios/tasks (i.e. different or unexpected data types). Specifically, genetic algorithmics and simulated annealing were implemented as the desired heuristics. To the best of the author knowledge, such an approach in the way it was developed has not yet been studied in literature.

\section{Methodology}

    This section intends to explain the coupling of meta-heuristics and artificial neural networks that was performed in our work. The benefits of the proposed technique are also overviewed, as they motivated its development.

    \subsection{Artificial Neural Networks}
    Artificial Neural Networks (ANN), a sub-topic of machine learning, have long been an interesting area of research, as they are closely modelled after the biological neural connections which make up the brain. Considering this, these networks are capable of almost mimicking the same functions that their biological counterparts perform, having varied applications in areas ranging from computer vision to content filtering to medical diagnostics. ANNs are organized by layers, receiving input and producing output after processing by a certain number of hidden layers. Each layer is made up of several neurons, with connections between these neurons at different layers depending on the network's internal structure. In general, basic training is composed of two processes: input forward propagation and error back-propagation. In the former, input is progressed in the network so as to generate an output by non-linear transformation and corresponding error, while in the latter the obtained error is distributed back, layer by layer, to each neuron. Through this process, layer weights and biases are adjusted minimizing the error with respect to the gradient direction.

    \subsection{Genetic Algorithms}
    These algorithms, detailed by Holland in \cite{meth_1}, represent a class of heuristic methods which, without prior training, aim to seek global solutions to a problem by simulating natural genetic selection and evolution. In this fashion, the process maintains a population of solutions at each iteration whose individuals (or chromosomes) are chosen based on their respective fittings to a chosen fitness function, typically probabilistic. Each chromosome is constituted by genes which characterize that particular solution, based on what values they take. These must belong to a set of possible values denominated alleles. Having chosen the members of the new population at each iteration they undergo genetic recombination, generating offspring and mutation. This way, natural selection is emulated and an increasingly fitting solution is obtained iteratively. Recombination and mutation methods are highly varied and application dependent, with some of the most common being described in \cite{meth_2}.
    
    \subsection{Simulated Annealing}
    This method, which was first introduced for combinatorial optimization in \cite{meth_3}, is motivated by the physical process of melting a solid and slowly decreasing its temperature so that it reaches a crystalline state of minimal energy while also avoiding undesirable meta-stable states \cite{meth_2}. As such, the technique is an adaptation of the Metropolis–Hastings algorithm in that a sequence of states (solutions) is obtained by continuously causing disturbances in the current state which has its own energy (fitness). At each iteration, a set of neighboring states to the current state are evaluated, one at a time. Should the energy of the new state be lower than that of the current state (meaning higher fitness), then the latter is replaced by the former as the current solution. In addition, if this does not occur, the new state may still replace the old one following a probability:
    
    \begin{equation}
        p = exp(\frac{E_i - E_j}{k_b \cdot T})
    \end{equation}{}
    
    Here $T$ denotes the temperature, $k_b$ is the Boltzmann constant, and the pair $(i,j)$ refers to the current and new states respectively, making $E_i$ and $E_j$ the corresponding energies.

    \subsection{Proposed Coupling}
    When considering each data instance has its own class label for supervised training, by looking at all instances together one is able to form a \textit{labeling code}, as a vector composed by the labels of all the considered data instances. This way, should an unlabelled dataset be composed of $L$ distinct instances each of which potentially being assigned to $C$ different classes, then a total of $C^L$ possible labelling codes can be formed for the dataset. Of course an evaluation of all potential label combinations is highly expensive in terms of computational power and can virtually last indeterminately. This motivates the use of metaheuristics to deal with this problem. Specifically, these methods have the ability to deal with the combinatorial explosion of unlabelled data, which can be particularly advantageous for deep learning problems such as this. Additionally, implementation of two inherently distinct heuristic methods may allow for a better understanding of how the solution search process is occurring, and consequently lead to better results. We performed this comparison between populational GA methodology and simulated annealing.
    
    \begin{figure}[!t]
        \centering
        \includegraphics[width=\columnwidth, height=12cm]{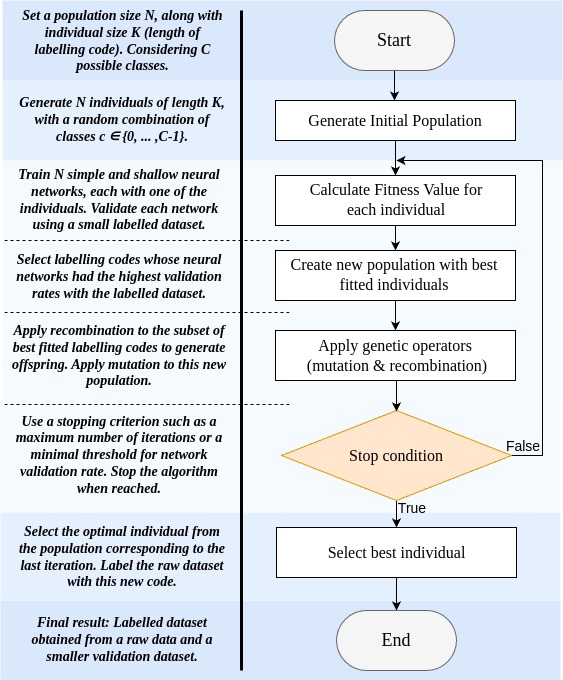}
        \caption{Flow diagram of standard genetic algorithmics (right), in parallel with the explanation of each step's specific functioning in our GA-ANN framework.}
        \label{fig:diagram}
    \end{figure}

    \begin{figure*}[ht!]
        \centering
        \includegraphics[width=0.99\textwidth]{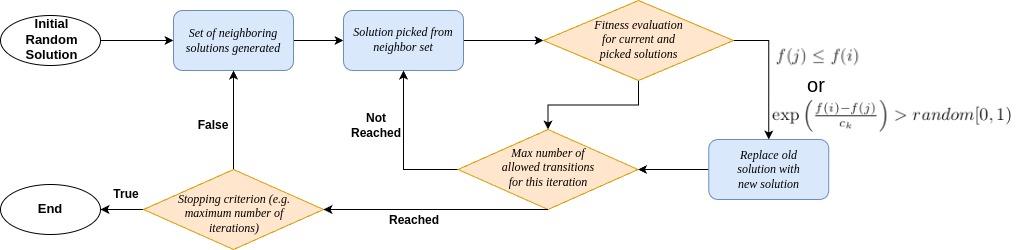}
        \caption{Flow diagram of the proposed Simulated Annealing technique.}
        \label{fig:diagram2}
    \end{figure*}
    
    The application of a heuristic such as a GA requires a cost function to evaluate the suitability of each considered solution. For the proposed technique, we employed the validation accuracy of trained neural networks. At each iteration of the GA, the current population is composed of $N$ individuals (i.e. $N$ distinct labelling codes for a dataset) each of which is used to train a neural model in parallel (i.e. all models have the same architecture). Evidently, given the potentially high number of networks to be trained, the chosen architecture should be simple and shallow, though hardware advancements may mitigate this issue. Using each individual of the population, the corresponding NN is trained and validated using a small set of ground truth data, labelled manually for instance. A group of the best fitted individuals (i.e. the ones whose models produce the highest validation accuracy), are chosen as the ``parents" of the next generation of individuals. To this group, recombination is applied to generate offspring individuals, followed by mutation applied to the entire population. Using a stopping criterion, this process is carried out for a number of iterations after which an optimal individual is picked from the population generated by the final iteration. A flow diagram of the proposed method is shown in Figure \ref{fig:diagram}, to ease the understanding of GAs as well as parallelly explain the specifics of each step in our framework. 
    
    As for the simulated annealing (SA) approach, the problem is identical since this technique is also able to deal with the combinatorial explosion of unlabelled data. Nevertheless, this issue is dealt with in a different manner since a solution is obtained and optimized by iteratively evaluating better options in its vicinity, instead of evaluating a population of solutions at each iteration. Given the method's nature, its implementation is explicit. Following the state machine depicted in Figure \ref{fig:diagram2}, an initial random solution is generated, for which a set of neighboring solutions is also obtained. Out of that set a new possible solution is picked and the two have their fitness evaluated. This evaluation is performed in the same manner as for the GA-ANN coupling previously described. As such, for the two solutions (labeling codes), two architecturally identical neural networks are instantiated, trained and validated against a small manually labelled set of data. As explained, should the fitness of the new labelling code meet the conditional or probabilistic criteria outlined, this will replace the old labelling code. This is repeated for a certain number of times at each iteration (number of allowed transitions), and for several iterations until a stopping criterion is reached. In terms of advantages, considering how only two neural networks must be developed at each iteration, this allows for much deeper and more complex models to be used as opposed to the ones employed in the GA. Thus, the SA approach allows for a more efficient use of the resources available, along with potentially better results. 
    
    Due to the randomness factor associated with the training of neural networks, the two techniques described above were also employed using conventional classifiers in lieu of the ANNs, for mere comparison of performances and efficiency. These included support vector machines (SVM) and random forests (RF).

\section{Experiments}

In order to test and validate the proposed techniques, the MNIST dataset for handwritten digit recognition \cite{deng2012mnist} was used. As it is representative of a typical supervised learning database, it is applicable to the development of automated data mining methods. In its current form, the dataset is composed of $60,000$ labelled pictures $28\times28$ of handwritten digits for training and an additional $10,000$ for testing. The pictures and corresponding labels are exactly as shown in Figure \ref{fig:codes}.
    
    \subsection{GA Method}
\begin{figure*}
        \centering
        \begin{subfigure}[t]{.5\textwidth}
          \centering
          \includegraphics[width=\linewidth]{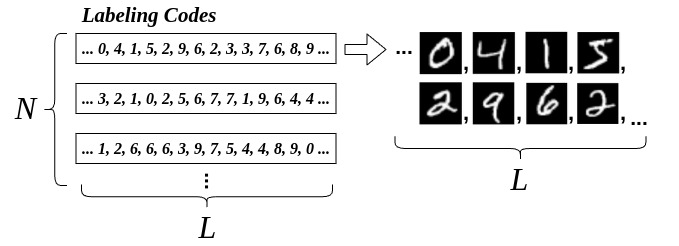}
          \caption{Labeling codes.}
          \label{fig:codes}
        \end{subfigure}%
        \begin{subfigure}[t]{0.5\textwidth}
          \centering
          \includegraphics[width=\linewidth]{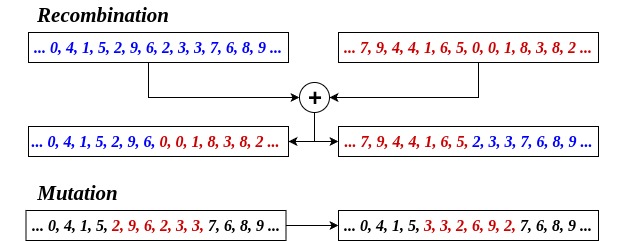}
          \caption{Recombination and mutation.}
          \label{fig:recmut}
        \end{subfigure}
        \caption{Examples for the labeling code vector format (left), used for testing with the MNIST dataset, as well as for the recombination and mutation methods (right), respectively performed as half-way merging and gene order inversion applied to the labelling codes.}
        \label{fig:2f}
    \end{figure*}

    In terms of the approach involving the combination of a genetic algorithm with neural networks or conventional classifiers, the recombination and mutation sections followed the diagrams shown in Figure \ref{fig:recmut}. In order to recombine two parent labeling codes, these were segmented in two and each two opposing segments, one of either parent, put together to form a new offspring. As for the mutation, a meere order inversion was performed on an adequate portion of the genes (individual labels) of each labeling code in the current solution population.

    For actual testing of the method, a small and well balanced random set of 200 images was obtained from the MNIST dataset. This smaller set was given a randomized initial labeling code for which the accuracy was measured against the images' ground truth. The population size was maintained at either 50 or 100 possible labeling codes. Further, the method was evaluated with 50, 100 and 150 iterations with the results being shown in the firt main column of Table \ref{tab:results}. Validation of each neural network or classifier integrated in the GA was performed each time using a distinct fixed size set of 45 images. All results shown were obtained by averaging those obtained from at least 5 runs of the same program, for increased robustness and confidence.
    
    
    Given the obtained results with the method, the exact same process was repeated, albeit with elitism incorporated into the technique. Simply put, at each iteration the two labeling codes providing the highest accuracies were maintained and automatically integrated in the population of the next iteration, without suffering mutation. The results of this second stage, which were also obtained after at least 5 runs of the same program, are shown in the second main column of Table \ref{tab:results}.
    

    \begin{table*}[!t]
\centering
\caption{Overall experimental results of combining ANN, SVM or RF with GA, without and with elitism (left and center), and SA (right). Highest validation accuracies are shown for the first and last iterations, at each column. }
\label{tab:results}
\resizebox{\textwidth}{!}{%
\begin{tabular}{@{}ccccccccccccccc@{}}
\cmidrule(l){4-15}
                      &                       &     & \multicolumn{4}{c}{GA w/o Elitism} & \multicolumn{4}{c}{GA w/ Elitism} & \multicolumn{4}{c}{SA} \\ \cmidrule(l){2-15} 
 &
  \multicolumn{2}{c}{Population Size} &
  \multicolumn{2}{c}{50} &
  \multicolumn{2}{c}{100} &
  \multicolumn{2}{c}{50} &
  \multicolumn{2}{c}{100} &
  \multicolumn{2}{c}{50} &
  \multicolumn{2}{c}{100} \\
 &
  \multicolumn{2}{c}{Accuracy {[}\%{]}} &
  Initial &
  Final &
  Initial &
  Final &
  Initial &
  Final &
  Initial &
  Final &
  Initial &
  Final &
  Initial &
  Final \\ \midrule
\multicolumn{1}{c|}{\multirow{3}{*}{NN}} &
  \multicolumn{1}{c|}{\multirow{9}{*}{Iterations}} &
  50 &
  \multirow{3}{*}{11.0} &
  12.7 &
  \multirow{3}{*}{13.5} &
  13.5 &
  \multirow{3}{*}{12.5} &
  20.0 &
  \multirow{3}{*}{13.0} &
  15.0 &
  \multirow{3}{*}{13.0} &
  8.2 &
  \multirow{3}{*}{9.4} &
  9.4 \\
\multicolumn{1}{c|}{} & \multicolumn{1}{c|}{} & 100 &       & 16.5     &      & 14.0     &      & 15.2     &      & 18.5     &    & 12.2  &   & 10.6  \\
\multicolumn{1}{c|}{} & \multicolumn{1}{c|}{} & 150 &       & 16.7     &      & 18.0     &      & 19.4     &      & 17.5     &    & 11.0  &   & 12.6  \\ \cmidrule(r){1-1} \cmidrule(l){3-15} 
\multicolumn{1}{c|}{\multirow{3}{*}{SVM}} &
  \multicolumn{1}{c|}{} &
  50 &
  \multirow{3}{*}{11.4} &
  16.2 &
  \multirow{3}{*}{13.4} &
  16.6 &
  \multirow{3}{*}{11.6} &
  17.0 &
  \multirow{3}{*}{12.6} &
  18.2 &
  \multirow{3}{*}{8.2} &
  7.8 &
  \multirow{3}{*}{9.4} &
  9.4 \\
\multicolumn{1}{c|}{} & \multicolumn{1}{c|}{} & 100 &       & 15.2     &      & 18.8     &      & 18.4     &      & 19.8     &    & 8.2   &   & 10.6  \\
\multicolumn{1}{c|}{} & \multicolumn{1}{c|}{} & 150 &       & 17.8     &      & 15.4     &      & 17.8     &      & 20.4     &    & 11.0  &   & 12.6  \\ \cmidrule(r){1-1} \cmidrule(l){3-15} 
\multicolumn{1}{c|}{\multirow{3}{*}{RF}} &
  \multicolumn{1}{c|}{} &
  50 &
  \multirow{3}{*}{11.4} &
  14.2 &
  \multirow{3}{*}{14.4} &
  12.6 &
  \multirow{3}{*}{11.5} &
  17.5 &
  \multirow{3}{*}{12.6} &
  17.0 &
  \multirow{3}{*}{13.2} &
  10.6 &
  \multirow{3}{*}{10.6} &
  12.2 \\
\multicolumn{1}{c|}{} & \multicolumn{1}{c|}{} & 100 &       & 16.2     &      & 15.8     &      & 19.7     &      & 17.0     &    & 8.8   &   & 10.4  \\
\multicolumn{1}{c|}{} & \multicolumn{1}{c|}{} & 150 &       & 15.4     &      & 16.6     &      & 16.7     &      & 17.8     &    & 9.2   &   & 11.0  \\ \bottomrule
\end{tabular}%
}
\end{table*}

    \subsection{SA Method}
    For the simulated annealing approach, used with the former for comparison with a non-populational technique, the same parameters were considered. This way, a similarly small and well balanced random set of 200 images was put together from the original MNIST dataset, as well as a 45 image validation set. The method was equally evaluated with 50, 100 and 150 iterations and the solution space for each iteration was set to have 50 or 100 possibilities. These possibilities were generated each time by resorting to random modification of a base solution to form new ones within the specified Mahalanobis distance \cite{mahalanobis}. The results of this experiment are shown in the third main column of Table \ref{tab:results}, having also been obtained after at least 5 runs of the same program.
    

\section{Discussion}

    As can be seen in Table \ref{tab:results}, results are similarly low, despite the genetic algorithm version giving somewhat better results than simulated annealing. In addition, a slight overall improvement of the results is also observed from the genetic algorithm with elitism, though the accuracies still not being high enough to be considered useful. As such it can be concluded that neither type of methods, populational or non-populational, were adequate for data mining attempts. Though there might be a common issue causing one or both methods to fail, it may also be the case that the techniques proposed are simply not suitable and other alternatives should be researched. In addition to the GA method performing better than its SA counterpart, small increases were also noted overall in the former by increasing the population size. Nevertheless, these still were not high enough to deem the technique a success. 
  
    Reasons for the lack of success of the proposed techniques may be many, though the most likely one seems to be the inadequacy of the fitness function employed. As it was empirically shown, a trained classifier's validation accuracy, using a ground truth dataset different from that used in training, is unsuitable for deciding whether or not the training dataset was correctly labeled and how the labels should be altered to increase the amount of correct ones. A reason for this may be the heterogeneity inherent to the data itself. For example, two different people may write the same digit very differently. Whilst it may be easier to recognize either handwriting representing the same digit, it is not as straightforward to validate the extrapolated characteristics that constitute a digit with knowledge obtained from only a few examples. This may be tackled by either homogenizing the datasets before employing the techniques on them, such as in \cite{Kilintzis_2024}, or by using generalization methods over extrapolated characteristics prior to the data mining stage. Another reason as to why the fitness function may have failed can be related with how classifiers are learning to classify. Despite doing it accordingly with the labels provided, classification may indicate the wrong digits, given their incorrectly constructed data representations, and then naturally flop their performance on the validation dataset, labeled only with correct ground truths. This results in the classifiers not being able to correlate the training and validations datasets, hence not optimizing the training data labels properly.

\section{Conclusions}

    In this paper, we proposed two methods to perform automated data mining through the combination of meta-heuristics with conventional classifiers and neural networks. These methods encompassed populational and non-populational approaches, as genetic algorithm and simulated annealing respectively. Both techniques were employed on a small subset of the MNIST dataset for handwritten digit recognition in order to assess their performance and ability to correctly label previously unseen data. As it was observed, the techniques faired rather unsuccessfully at the data mining task, highly likely due to the inadequacy of a ground truth dataset validation accuracy as a fitness function. This unsuitability may have been related with issues stemming from data heterogeneity or incorrect, though congruent, feature learning by the classifiers and neural networks
    
    In the future, we intend to tackle the fitness function issue by attempting to further generalize the extrapolated data features either by employing deeper networks or through other probabilistic methods. With this it is expected that the techniques will improve their performance and be able to successfully perform automated data mining independent of its type.



\balance

\bibliographystyle{ieeetr}
\bibliography{biblio}

\end{document}